\documentclass[journal,twoside,web]{ieeecolormod}
\usepackage{genericmod}
\usepackage{amsmath,amssymb,amsfonts}
\usepackage{algorithmic}
\usepackage{graphicx}
\usepackage{algorithm,algorithmic}
\usepackage{hyperref}
\hypersetup{hidelinks}
\usepackage{textcomp}
\usepackage{booktabs} 
\usepackage{multirow}
\usepackage{placeins}
\usepackage{xcolor}
\usepackage[absolute, overlay]{textpos}

\usepackage{cite}  
\def\BibTeX{{\rm B\kern-.05em{\sc i\kern-.025em b}\kern-.08em
    T\kern-.1667em\lower.7ex\hbox{E}\kern-.125emX}}
\begin{document}
\title{Generating Completions for Broca's Aphasic Sentences Using  Large Language Models}
\author{Sijbren van Vaals, Yevgen Matusevych, Frank Tsiwah 
\thanks{This work was (partly) supported by the PPS grant of the Ministry of Economic Affairs and Climate Policy through CLICKNL, the Top Consortium for Knowledge and Innovation (TKI) in the Creative Industries. We would like to acknowledge the support of AphasiaBank for providing us with the data used for this study. (Corresponding author: Frank Tsiwah, f.tsiwah@rug.nl).} 
\thanks{All authors are with the Center for Language and Cognition Groningen, University of Groningen, Oude Kijk in ’t Jatstraat, 26, 9712 EK, Groningen, the Netherlands (e-mails: s.j.van.vaals@rug.nl, yevgen.matusevych@rug.nl, f.tsiwah@rug.nl).}
}

\begin{textblock*}{\textwidth}(1.7cm, 0.5cm)
{\color{red}\small \noindent © 2025 IEEE. Personal use of this material is permitted. Permission from IEEE must be obtained for all other uses, in any current or future media, including reprinting/republishing this material for advertising or promotional purposes, creating new collective works, for resale or redistribution to servers or lists, or reuse of any copyrighted component of this work in other works. The final published version is available at \url{https://doi.org/10.1109/jbhi.2025.3639109}}
\end{textblock*}

\maketitle
\thispagestyle{empty}

\markboth{}{Completing Broca's aphasic sentences with LLMs}

\begin{abstract}
Broca's aphasia is a type of aphasia characterized by non-fluent, effortful and agrammatic speech production with relatively good comprehension. Since traditional aphasia treatment methods are often time-consuming, labour-intensive, and do not reflect real-world conversations, applying natural language processing based approaches such as Large Language Models (LLMs) could potentially contribute to improving existing treatment approaches. To address this issue, we explore the use of sequence-to-sequence LLMs for completing Broca's aphasic sentences. We first generate synthetic Broca's aphasic data using a rule-based system designed to mirror the linguistic characteristics of Broca's aphasic speech. Using this synthetic data (without authentic aphasic samples), we then fine-tune four pre-trained LLMs on the task of completing agrammatic sentences. We evaluate our fine-tuned models on both synthetic and authentic Broca's aphasic data. We demonstrate LLMs' capability for reconstructing agrammatic sentences, with the models showing improved performance with longer input utterances. Our result highlights the LLMs' potential in advancing communication aids for individuals with Broca's aphasia and possibly other clinical populations.

\end{abstract}


\section{Introduction and Background} \label{sec:IntroBack}
\pagenumbering{arabic}

Aphasia is an acquired language disorder after a focal brain injury. Typically, aphasia has long-term consequences on the communication abilities of the affected individuals, and thus, negatively impacts their quality of life \cite{Berthier2005, Hilari2010} together with their relatives' quality of life \cite{azevedo2023understanding}. Within the spectrum of aphasia, Broca's aphasia, the target group for this study, is a sub-type of aphasia described as having non-fluent, effortful, telegraphic speech production, but with relatively preserved comprehension. Given the significant language fragmentation in Broca's aphasic speech, diverse treatment approaches have been proposed \cite{Cuperus2023-jz}. While studies have demonstrated the effectiveness of these treatments, mixed findings regarding long-term maintenance of gains post-treatment \cite{Nickels2002, deAguiar2016} have led to the emergence of alternative communication aids. These alternative approaches, although beneficial, present some limitations such as inefficiency, lack of personalization, and poor generalization to real-world conversations \cite{azevedo2023understanding}. Existing communication aids for people with aphasia (PWA) include low-tech options like picture or spelling boards and high-tech devices with pre-defined messages (such as making an order for coffee), but these tools often limit expression to basic ideas or messages and can be challenging for PWA to navigate \cite{Koul2005, Koul2008, Mooney2018, Alam2023, Chavers2021, azevedo2023understanding}. Furthermore, these alternative communication aids frequently require specific skills, such as keyboard use or navigating pictograms, which can pose significant challenges for many PWA, thereby restricting their ability to engage in daily conversations \cite{azevedo2023understanding}. Navigating pictogram sets or relying on pre-defined texts often does not fully capture the contextual dependencies inherent in real-world conversations. Consequently, developing interventions that can replicate the fluid, context-dependent language patterns observed in spontaneous interactions with familiar partners, such as spouses and caregivers, is of considerable importance. The goal of the current study is to investigate the potential of reconstructing agrammatic language into full utterances by using Large Language Models (LLMs) fine-tuned on synthetic Broca's aphasic data\footnote{The code and data are available online: \url{https://github.com/sijbrenvv/Completions_for_Broca-s_aphasia}}.
 
\subsection{Linguistic characteristics of Broca's aphasic language}

Broca's aphasia is an acquired language disorder typically resulting from brain injury (e.g., stroke) or brain degeneration, which impacts an individual's ability to produce spoken and written language, with comprehension remaining relatively intact \cite{thompson1995analysis, bastiaanse2012perspectives,perez2023seq2seq,declercq2024detecting}. 
Broca's aphasic spontaneous speech is characterized by several linguistic deficits such as omission of bound morphemes, function words (especially determiners), difficulty in producing verbs with complex argument structure, and impairment in complex syntactic structures \cite{Shapiro1987, Shapiro1993, thompson1995analysis,goodglass1983assessment,grodzinsky1984syntactic,SAFFRAN1989440,goodglass2001bdae,kolk2006language,ruiter2013combining}. Verbs with one argument tend to be relatively easier to produce than verbs with two or more arguments \cite{Shapiro1987, thompson1997patterns}.
Content words generally tend to be preserved.

Despite the difficulties in production, comprehension abilities in Broca's aphasia remain relatively unaffected. This discrepancy suggests that linguistic representations may be preserved in PWA, but access to these representations is impaired \cite{Shapiro1987, SAFFRAN1989440, Shapiro1993}. In summary, individuals with Broca's aphasia produce highly agrammatic speech, with one- to two-word utterances in more severe cases.

\subsection{AI applications in aphasiology}
In recent years, Natural Language Processing (NLP) techniques have been adopted in aphasia research as well as clinical applications, particularly in analyzing connected speech production by PWA \cite{themistocleous2021automatic,fraser-etal-2019-importance, perez2023seq2seq, declercq2024detecting}. While traditional manual analysis of patients' speech transcription is time-consuming and labor-intensive, automated NLP approaches have significantly streamlined this process \cite{perez2023seq2seq}. Applying NLP methods not only saves time but also provides consistent and accurate evaluations, potentially aiding in the diagnosis and classification of aphasia subtypes \cite{themistocleous2021automatic}, which could ultimately inform more targeted treatment strategies. 

Previous research has primarily focused on using NLP techniques for the automatic detection of aphasia (with potential application in diagnostic purposes), while largely overlooking the use of these techniques for developing assistive technologies to enhance communication for PWA, their spouses and caregivers. Recent scoping reviews underscore significant gaps in the application of AI within aphasiology, particularly in its therapeutic and assistive uses \cite{Adikari2024, Privitera2024, Azevedo2024}. One review \cite{Adikari2024} reported that 7 out of their 77 reviewed studies explored applications of AI in therapeutic or assistive applications, while most research focused on automated assessment and diagnosis of aphasia. The review highlighted a predominant reliance on supervised machine learning for classification tasks, with limited use of advanced AI techniques for personalized rehabilitation or self-management. Similarly, \cite{Azevedo2024} did not identify any studies integrating AI into augmentative and alternative communication (AAC) devices specifically tailored for aphasia rehabilitation. Together, these findings emphasize a disproportionate focus on diagnostic applications over functional communication support, revealing a critical need for AI-driven innovations that address real-world conversational challenges faced by PWA and their caregivers.

At the same time, \cite{misra2022assistive} recently reported an innovative approach that could potentially contribute to the advancement of assistive technology by using LLMs for reconstructing agrammatic aphasic sentences.
The goal of that study was twofold. First, LLMs require large amounts of training data, while large datasets of aphasic speech do not exist. Therefore, the study \cite{misra2022assistive} proposes a system that takes as an input a corpus of written texts from healthy speakers (C4 dataset \cite{raffel2020t5}), and generates synthetic aphasic utterances by modifying the sentences in the corpus so that their linguistic characteristics match those reported for Broca's aphasic speech. Second, the authors then use this synthetic dataset to fine-tune a pre-trained LLM, specifically the T5 model \cite{raffel2020t5}, on a sentence completion task. Although the study \cite{misra2022assistive} showed promising results, it presents a few methodological problems. First, the written text in the C4 corpus (used to generate synthetic aphasic utterances) does not accurately represent daily spoken language, potentially limiting the real-life applicability of the resulting system. Second, even though the study \cite{misra2022assistive} aims to demonstrate the potential of synthetic datasets for developing communication aids for PWA, the lack of testing on real aphasic utterances makes it unclear to what extent the presented approach is applicable to actual clinical scenarios. Lastly, the study's use of the C4 dataset for generating synthetic aphasic sentences and evaluating the fine-tuned LLM causes a potentially serious issue of data leakage, since the C4 dataset was included in the original training data for the T5 LLM. A data leakage could lead to artificially inflated performance metrics and even compromise the validity of the model's evaluation and its future application on real aphasic speech.

The present study aims to address these problems by answering the following research question: To what extent can a pretrained sequence-to-sequence LLM, fine-tuned on synthetic Broca's aphasic data, effectively reconstruct authentic agrammatic sentences produced by individuals with Broca's aphasia into complete, grammatically correct sentences?  To address this research question, we first generate synthetic Broca's aphasic sentences based on spoken language sentences produced by neurotypical individuals. This is done using a rule-based system adapted from the above-mentioned study \cite{misra2022assistive}.  Second, we use our synthetic data to fine-tune a series of pretrained LLMs on a sentence completion task, and then evaluate these LLMs on synthetic and authentic aphasic sentences.

\section{Method} \label{sec:Method}

We aim to assess the extent to which fragmented Broca's aphasic sentences can be reconstructed using sequence-to-sequence LLMs fine-tuned on synthetic aphasic data.
Our overall methodological setup, therefore, consists of two steps: (1) synthetic data generation, and (2) fine-tuning LLMs on a sentence completion task using synthetic data. 

For the synthetic data generation, we employ a tailor-made rule-based system inspired by the approach of \cite{misra2022assistive}. Its rules are based on the linguistic features associated with Broca's aphasic speech, deduced from existing literature. This approach ensures that our synthetic data resembles Broca's aphasic speech in terms of its linguistic characteristics.

For the sentence completion task, we fine-tune four sequence-to-sequence LLMs from the T5 \cite{raffel2020t5} and \textsc{Flan}-T5 families \cite{chung2022flant5} on the task of completing synthetic Broca's aphasic sentences generated by our rule-based system. We quantitatively evaluate the resulting models and then test 
the best-performing model on authentic Broca's aphasic sentences. Since the authentic sentences do not have ground-truth data (i.e., target completions), in this case we only assess the generated completions qualitatively.

\subsection{Synthetic data generation} 
\label{sec:syngen}

To generate synthetic sentences, we use an adapted version of the general approach described in \cite{misra2022assistive}. The system takes a natural language corpus and, using a set of predefined rules, modifies sentences so that they resemble fragmented sentences commonly produced by PWA. This approach effectively transforms each original sentence into a single synthetic aphasic sentence. Compared to the original system of \cite{misra2022assistive}, we use different sentence modification rates, additional features, and rules to reduce the complexity and grammaticality of sentences.

First, we consider a natural language corpus (see Section~\ref{sec:Data} below for more details) and discard all sentences which are longer than 15 words or, following \cite{misra2022assistive}, contain punctuation (other than a comma or full stop) or special characters. We also extract verb and noun phrases using spaCy \cite{spacy} and discard sentences with the ratio of noun phrases to verb phrases greater than $2$, based on existing studies of aphasic speech \cite{thompson1995analysis, tetzloff2018quantitative}.

For the retained sentences, we use a spaCy pipeline \cite{spacy} with the UDPipe English model \cite{straka-etal-2016-udpipe} to obtain a dependency parse for each sentence and retrieve each word's part-of-speech (PoS) tag. We then probabilistically apply a series of PoS-specific rules to each word (see Table~\ref{tab:rules}), with probabilities established based on \cite{misra2022assistive} and our manual inspection of generated sentences.
Table \ref{tab:syn_data} shows examples of synthetic utterances next to their corresponding original utterances, and highlights how the synthetic data retains the core content of the original sentences while exhibiting the fragmented and simplified structure typical of Broca's aphasic speech.

\begin{table}[h]
\caption{Rules used by our synthetic data generation system.}
\label{tab:rules}
\begin{tabular}{p{1.7cm}p{3.5cm}p{0.8cm}p{1.2cm}}
\toprule
\textbf{POS tag}                     & \textbf{Rule}                                                                                                    & \textbf{Prob.} & \textbf{Example}   \\ \midrule
Noun                                 & Change grammatical number & 30 \%                & \textit{table $\rightarrow$ tables} \\ \midrule
Adjective / \newline adverb / verb                   & Discard                                                                                                          & 50 \%                & \textit{regular $\rightarrow \varnothing$ } \\ \midrule
Pronoun (possessive / \newline demonstrative) & Replace with another pronoun of the same type                                               & 40 \%                & \textit{this $\rightarrow$ those} \\ \midrule
Determiner / \newline adposition / \newline particle   & Discard & 70 \%                & \textit{the $\rightarrow \varnothing$} \\ \midrule
Other                                & Do not change                                                                                                             & 100 \%               & N/A                \\ \bottomrule
\end{tabular}
\end{table}

\begin{table}[t]
\caption{Examples of synthetic utterances generated by the rule-base system.}
\label{tab:syn_data}
\begin{tabular}{p{5cm}p{3cm}}
\toprule
\textbf{Original utterance} & \textbf{Synthetic utterance} \\ \midrule
But we are talking just the regular light horses you know. & But we are just light horse you know. \\ 
\midrule
That band had sung that very night. & Band have very night. \\ 
\midrule
Which just does not sound like a very practical proposal. & Which do sound a very proposal. \\
\midrule
I mean crumbed and in the jar. & I crumb and the jar. \\ \bottomrule

\end{tabular}
\end{table}

In some extreme cases, a simultaneous application of multiple rules could result in having unrealistically short utterances. Therefore, as a final step, we exclude synthetic utterances that do not adhere to either or both of the following length requirements: (1) the synthetic utterance must have at least three words, and (2) the number of words in a synthetic utterance must be within the range $25$--$75 \%$  of the number of words in the respective original utterance.

Ideally, our synthetic sentences should be similar to authentic aphasic sentences and different from sentences produced by neurotypical individuals. There are many possible ways to compare sentence characteristics, but what matters in our setup is that the patterns described above hold in the context of LLMs. This is why we estimate the average acceptability (or predictability) of a sample of sentences in each data set using two language models commonly used for this purpose, GPT-2 \cite{radford2019language} and RoBERTa \cite{liu2019roberta}. Predictability is estimated through the standard measure called surprisal, or negative log-probability \cite{levy2008expectation, goodkind2018predictive}, which measures how easy it is on average for a language model to predict words in a sentence given their context. We apply this approach to sentences produced by neurotypical individuals, PWA, and our synthetic data generation system. We only consider sentences that are longer than 3 words and downsample each data set to 4,448 sentences (the size of the smallest, authentic aphasic, data set after filtering out short sentences). We expect that the language produced by neurotypical individuals should be highly predictable and, therefore, associated with lower surprisal, while both authentic and synthetic authentic data should be more irregular, and, therefore, less predictable and associated with higher surprisal.

The presented approach relies on two types of data. To produce a number of synthetic utterances sufficient for fine-tuning a LLM, our system needs a large set of utterances produced by neurotypical individuals. To evaluate our system, we additionally need a set of authentic utterances produced by PWA. We describe the relevant data sources in Section~\ref{sec:Data} below, but first we present our sentence completion models.

\subsection{Sentence completion}

We use a synthetic dataset obtained as described above to fine-tune sequence-to-sequence LLMs on the task of sentence completion. We choose four large pretrained encoder-decoder models based on their performance in common NLP tasks, the availability of their source code, the computational resources available to us, and, for T5, its use in an earlier study that is most similar to ours \cite{misra2022assistive}. Each LLM has been pretrained on large amounts of text data, and in our setup we only fine-tune it on our target task. 

\subsubsection{Models} 

\textsc{T5-base} and \textsc{T5-large} \cite{raffel2020t5} belong to the Text-to-Text Transfer Transformer (T5) family. These models are based on the transformer architecture \cite{vaswani2017attention} and leverage transfer learning through the use of task-specific prefixes that allow framing any NLP problem as a text-to-text task. The models consist of encoder and decoder components with stacked layers, self-attention mechanisms, and feed-forward networks. There are several versions of the model that differ in the number of parameters, and we use two of them: \textsc{T5-base} is the baseline version of the model with 220M parameters, and \textsc{T5-large} has 770M parameters. While there are even larger models in the T5 family, we could not use them in this study due to the lack of appropriate computational resources. All T5 models have been pretrained on the large Colossal Clean Crawled Corpus (C4) dataset, comprising 750~GB of text crawled from the web.

\textsc{\textsc{Flan}-T5-base} and \textsc{\textsc{Flan}-T5-XL} belong to the \textsc{Flan}-T5 family \cite{chung2022flant5}. These models are based on the T5 models but are extended using instruction fine-tuning on a large variety of tasks (dialog, arithmetic reasoning, etc.), which improves the models' generalization to unseen tasks. \textsc{\textsc{Flan}-T5-base} has 250M parameters, while \textsc{\textsc{Flan}-T5-XL} has 3B parameters. Again, we could not use larger versions of the \textsc{Flan} models due to our computational limitations. 

\subsubsection{Fine-tuning setup}

We fine-tune each of the four models on a synthetic aphasic dataset generated as described in the previous section. During this process, the models learn to reconstruct (synthetic) fragmented Broca's aphasic sentences. We use ChrF (CHaRacter-level F-score \cite{popovic-2015-chrf}) as the evaluation metric for updating the models' parameters. To take into account possible influence of prefixes on the models' performance, we fine-tune the models in two conditions: without any prefix and with prefix ``Complete this sentence'' (for comparison to \cite{misra2022assistive}).

We use the HuggingFace transformer library \cite{wolf-etal-2020-transformers} to load and quantise each model to 4-bit precision with a batch size of 8. For efficiency, we employ HuggingFace's Parameter-Efficient Fine-Tuning (PEFT) library \cite{peft} that incorporates Quantized Low-Rank Adaptation (QLoRA) \cite{dettmers2023qloraefficientfinetuningquantized} (see more details in Supplementary Materials). 
Using these techniques, we fine-tune each model. \textsc{T5-base} is fine-tuned for a maximum of 5 epochs with early stopping after 2 epochs with no improvement, using a learning rate of $10^{-4}$ with half precision (FP16). The other models are fine-tuned for $1$ epoch only due to overfitting. Supplementary Materials provide technical details about fine-tuning each model.

To evaluate our sentence completion models, we use two different methods. First, we compare the models' completions of (unseen) synthetic sentences to their original counterparts (i.e., the same sentences before they are processed by our synthetic data generation system).
This comparison is done using three measures: ChrF and RougeL \cite{lin-2004-rouge} between the two sentences (mentioned above), and cosine similarity between the embeddings of the two sentences (using embeddings of \cite{ni2021sentencet5}). While ChrF and RougeL measure the degree of surface-level (i.e., character and word, respectively) alignment between the two sentences, cosine similarity measures how close the meanings of the two sentences are. Although BLEU \cite{papineni2002bleu} is a popular evaluation metric in sentence comparison tasks, we found it to be unsuitable for our sentence completion task. The explanation with examples can be found in Supplementary Materials. Second, we use our models to generate completions for authentic sentences produced by PWA. In this case, we have no ground-truth data to compare our models' completions to, and we limit our evaluation to qualitative analyses.

\subsection{Data sources}
\label{sec:Data}

Considering the setup described in the two previous sections, we need two main types of data: (1) authentic utterances produced by PWA, to be used for the evaluation of our synthetic data generation system and our sentence completion models, and (2) utterances produced by neurotypical individuals, to be used as input to our synthetic data generation system, as well as for its evaluation. Ideally, the two types of data should be similar in their register or even content. We use the following data sources.

\subsubsection{Authentic aphasic data} To obtain authentic data produced by PWA, we use AphasiaBank \cite{macwhinney2011aphasiabank}, a foundational resource in aphasia research containing interactions between PWA and clinicians. It provides speech data collected from PWA during clinical studies. 
We use this data for the final evaluation of our sentence completion models, where the models generate completions for these utterances.

\subsubsection{Neurotypical individuals' data from AphasiaBank} AphasiaBank also contains utterances produced by neurotypical individuals (i.e., the control condition in many clinical studies) in the same setting as our authentic aphasic data. Thanks to their close similarity to the aphasic data, these utterances perfectly suit our goal of evaluating the quality of our synthetic Broca's aphasic utterances. At the same time, we decided against using this data as an input to our sentence generation system, because the content of these utterances can be very similar to that of the authentic aphasic utterances: for example, both neurotypical individuals and PWA could be describing the same picture as part of the data collection procedure. Therefore, feeding the neurotypical utterances from AphasiaBank into our sentence generation system would yield synthetic data that's potentially too similar to our evaluation data, leading to an unfair evaluation of our sentence completion models.

\subsubsection{Neurotypical individuals' data from SBCSAE}
The Santa Barbara Corpus of Spoken American English (SBCSAE) \cite{dubois2000santa} is a compilation of naturally occurring spoken interactions containing approximately 250k word tokens. SBCSAE is comparable to AphasiaBank in that it consists of transcriptions of spoken language. Finally, the corpus is very diverse and representative, as the data was collected from individuals of varied regional origin, age, ethnic and social background, occupation, and gender. We use utterances from SBCSAE to provide input to our synthetic data generation system, as well as to evaluate this system. 

The sentences in both AphasiaBank and SBCSAE include linguistic annotations, special characters, and other types of noise. We preprocess the data sources to extract only the actual word tokens and remove all annotations (i.e., pauses, actions, text within angle and square brackets), unicode errors and special characters, instances of stuttering and repeated words, etc. A full list of preprocessing routines with examples can be found in Supplementary Materials.

\section{Results}
\label{sec:Results}

In this section, we first evaluate our synthetic data generation system. We then assess the LLMs' ability to reconstruct synthetic Broca's aphasic sentences.
Finally, we evaluate the best-performing LLM on completing authentic sentences produced by PWA.

\subsection{Evaluating synthetic data generation system} \label{sec:quality}

\begin{table}
\centering
\caption{Average sentence predictability (surprisal) assigned to each data type by two pretrained LLMs. While the synthetic aphasic sentences are not directly extracted from any specific source, they are generated based on the SBCSAE data.}
\label{tab:data_quality}
\begin{tabular}{llrr}
\toprule
\textbf{Data type} & \textbf{Source} & \multicolumn{1}{l}{\textbf{GPT-2}} & \multicolumn{1}{l}{\textbf{RoBERTa}} \\ \midrule
Neurotypical       & SBCSAE          & 4.76                               & 2.96                                 \\
Neurotypical       & AphasiaBank     & 5.43                               & 3.50                                  \\
Authentic aphasic  & AphasiaBank     & 6.48                               & 5.08                                 \\
Synthetic aphasic  & -- (SBCSAE)     & 6.39                               & 5.51                                 \\ \midrule
\end{tabular}
\end{table}

To evaluate our synthetic data generation system, we consider the average sentence predictability (surprisal) values shown in Table~\ref{tab:data_quality}. Note that GPT-2 and RoBERTa have very different architectures and are trained on different data sets, therefore, the surprisal values \textit{across} the two models (columns) are not directly comparable, and we only compare the values within each model (column). We first notice that the two data sets that contain sentences from neurotypical individuals (rows 1--2) are associated with lower surprisal values compared to the authentic aphasic data (row 3), and this result holds for the values generated by both language models, the pattern we expected. Second, our synthetic data (row 4) yields surprisal values that are comparable to the values for the authentic aphasic data (row 3) and higher than those for the neurotypical data (rows 1--2). This is an important result that suggests that our synthetic generation system produces sentences where words are harder for LLMs to predict from context -- similar to authentic aphasic sentences and in contrast to neurotypical language.
Upon manual inspection, our generated data also appears satisfactory, therefore, we proceed with using the synthetic Broca's aphasic data generated by our system to fine-tune the four LLMs.

\subsection{Evaluating sentence completion models} \label{sec:syn_comp}

\subsubsection{Synthetic data} 

In this setting, we compare the agrammatic (synthetic) sentences completed by our four LLMs to the original sentences.

\begin{table}
\centering
\caption{Evaluation results of the sentence completion models on the synthetic test set. Mean and standard error are provided for each measure.}
\label{tab:model_stats}
\resizebox{\columnwidth}{!}{
\begin{tabular}{lcccccc}
\toprule
& \multicolumn{3}{c}{\textbf{Without prefix}} & \multicolumn{3}{c}{\textbf{With prefix}} \\
\cmidrule(llr){2-4} \cmidrule(llr){5-7}
\textbf{Model}         & \textbf{ChrF} & \textbf{RougeL}  & \textbf{Cos. sim.}  & \textbf{ChrF} & \textbf{RougeL}    & \textbf{Cos. sim.} \\
\midrule
\textsc{T5-base}            & 56.39$\pm$0.48  & 0.71$\pm$0.00  & 0.88$\pm$0.00           & 56.56$\pm$0.48  & 0.71$\pm$0.00       & 0.88$\pm$0.00           \\
\textsc{Flan-T5-base}        & 49.53$\pm$0.42  & 0.67$\pm$0.00      & 0.87$\pm$0.00           & 50.15$\pm$0.42  & 0.67$\pm$0.00          & 0.87$\pm$0.00          \\
\textsc{T5-large}            & 53.95$\pm$0.46  & 0.68$\pm$0.00     & 0.88$\pm$0.00           & 54.94$\pm$0.47   & 0.70$\pm$0.00       & 0.88$\pm$0.00          \\
\textsc{Flan-T5-XL}          & 55.71$\pm$0.47  & 0.69$\pm$0.00     & 0.88$\pm$0.00            & 56.19$\pm$0.48   & 0.69$\pm$0.00      & 0.88$\pm$0.00           \\
\bottomrule
\end{tabular}
}
\end{table}

\textsc{T5-base} model demonstrates consistent performance on the synthetic test set regardless of whether the prefix ``Complete this sentence'' is used. The other models show slightly improved performance with the prefix, which suggests that they may benefit from additional contextual cues provided by the prefix.

A visual inspection of the generated completions (see examples in Table~\ref{tab:syn_comp}) reveals that the models, especially \textsc{T5-base} and \textsc{Flan}-T5-XL, show great capacity in addressing some of the main deficits observed in Broca's aphasic utterances, such as word omission.
While \textsc{T5-base} achieves the highest performance scores, it occasionally generates semantically incorrect completions, which highlights limitations in its ability to generate contextually appropriate sentences. 
\textsc{Flan}-T5-XL, on the other hand,  performs consistently well, often opting for simpler but semantically plausible completions. 
This tendency reveals a trade-off between complexity and fluency, and suggests that reconstructing more morphosyntactically complex utterances is a challenging task for our models. 
More examples of generated completions for the synthetic test set are presented in Supplementary Materials.

\begin{table}
\caption{Examples of LLMs' completions for synthetic utterances. Prefix used in all cases.}
\label{tab:syn_comp}
\begin{tabular}{lp{2.5cm}p{3.5cm}}
\toprule
\textbf{Model}    & \textbf{Synthetic utterance} & \textbf{Generated completion}     \\ \midrule
\multirow{8}{*}{\textsc{T5-base}}     & I do i thethey they did   & I do not know i do not know thethey did not know.  \\ 
& I voting favor of it.     & I am voting in favor of it.  \\ 
& I said you me better than i myself.   & I said you would know me better than i did myself.  \\  
& Lenore what it.           & Lenore knows what it is.   \\  
& They our lawns.           & They are going to have to cut their lawn. \\ \midrule 
\multirow{5}{*}{\textsc{Flan}-T5-XL}  & Big or go all.            & Big or go all the way.  \\ 
& No you it.                & No you can not do it.  \\ 
& I will worry it.   & I will not worry about it.   \\ 
& I will him call.          & I will call him up.   \\ 
& It lot first.             & It was a lot of fun at first. \\ \bottomrule
\end{tabular}
\end{table}

\begin{table*}
\centering
\caption{Examples of LLMs' completions for all authentic utterances (top half) and with  one-word and two-word utterances excluded (bottom half). The prefix is used in all cases.}
\label{tab:auth_comp}
\begin{tabular}{p{5cm}p{6cm}p{6cm}}
\toprule
\textbf{Authentic aphasic utterance}   & \textbf{Generated by \textsc{T5-base}}     & \textbf{Generated by \textsc{Flan}-T5-XL}   \\
\midrule
Two slice.        & Two slices.      & Two slices.   \\                  
Oh ohman.         & Oh ohman.      &  Oh ohman.  \\ 
Hold on a second.  & Hold on for a second.      &  Hold on a second.  \\ 
Peanut , bæ@you ,.  & Peanut, b@you,.      & Peanut, b@you,.  \\ 
So it is youknow cinderella beautiful person youknow.    & So it is you know cinderella is a beautiful person youknow.      &  So it is youknow cinderella beautiful person youknow.  \\ 
Because i it nothing.  & Because i thought it was nothing.      &   Because i do not know it nothing. \\ 
Everything's okay. & Everything's okay.   &  Everything is okay.    \\ 
Because kid in here.   &  Because there is a kid in here.   &  Because the kid is in here.  \\ 

So that is.       & So that is.   &  So that is what it is.    \\ 
Nice !.           & Nice!   &  Nice!.   \\ 
And ride away because bebecause cinderella was good.  & And then ride away because bebecause cinderella was good.   &  And ride away because bebecause cinderella was good.   \\
\midrule
What in the world is this ?. & What in the world is this? & What in the world is this? \\ 
And he is.                   & And he is. & And he is not. \\ 
Oh is it hid feet.           & Oh is it hid under the feet. & Oh is it hid under my feet. \\ 
And peanut butter is.        & And peanut butter is not. & And peanut butter is. \\ 
I do it.                     & I do not know how to do it. & I do not do it. \\ 
You hafta come with these.   & You hafta come with this. & You hafta come with these. \\ 
It was old time.             & It was old time. & It was an old time. \\ 
And this is with them.       & And this is what is with them. & And this is what is going on with them. \\ 
Ah i have lost my.           & Ah i have lost my. & Ah i have lost my mind. \\ 
I know say.                  & I know say. & I know what to say. \\ 
And you just.                & And you just. & And you just just sat there. \\ \bottomrule
\end{tabular}
\end{table*}

\subsubsection{Authentic data}

In this setting, we have no ground-truth data for authentic Broca's aphasic utterances. Therefore, for these utterances, we only present a qualitative analysis of sentence completions generated by \textsc{T5-base} and \textsc{Flan}-T5-XL fine-tuned using the prefix ``Complete this sentence''. We decided to focus on these two models as they achieved the highest performance on the synthetic data.

Table~\ref{tab:auth_comp} (top half) presents examples of completions generated by the two LLMs.
Although the models exhibit some capability of handling Broca's aphasic speech, important limitations remain. In particular, both models sometimes only minimally change the authentic utterance or even simply reproduce it.
Also, they occasionally add negations (2.99\% of the time for FLAN-T5-XL), which can change the sentence meaning to the opposite, or expand contractions wrongly (e.g., ``He's'' to ``He is'' instead of ``He has''). 
The occasional addition of negation reflects the general difficulty of this phenomenon for natural language processing models \cite{Hosseini2021UnderstandingBU,Rezaei2024ParaphrasingIA}.

While there are instances of grammatical completions, the models often produce outputs that do not correct the grammatical errors present in the authentic Broca's aphasic utterances. This indicates a gap in the models' ability to fully interpret and rectify the Broca's aphasic speech patterns. Both models occasionally insert common conversational fillers or affirmations such as ``Yeah'', ``Okay'' and ``I do not know'', which do not meaningfully contribute to completing the utterance but reflect an attempt to provide a plausible conversational continuation.
There are some utterances so agrammatic that completing them meaningfully would be challenging even for humans. In these cases, the models either do not change the utterances at all or provide completions that do not improve clarity. For example, ``Peanut, b@you,.'' by \textsc{T5-base} remains largely unreadable.

To estimate how well FLAN-T5-XL performs in the target task, we have manually annotated 10\% of our test data, namely 455 sentences. We did the annotation without considering their broader context, to simulate the task that the model was faced with. We found that 196 sentences did not require any completion by the model, while 259 did. From the sentences that needed to be completed, 66 would not be possible to complete, given the lack of the broader context, while 193 could in principle be completed. Out of these, the model has successfully completed 80 sentences, simply reproduced 57 of them, and produced errors in 56 sentences (42 cases of incomplete/minimal change, 12 unnecessary additions, and 2 cases of added negation). On top of that, the model introduced unnecessary changes into the 13 (out of the 196) sentences that did not require completion.

Very short utterances might be difficult for our models to complete not only because they are often cryptic, but also because our models were not fine-tuned on this task: recall from Section~\ref{sec:syngen} above that our synthetic utterances had to contain at least three words. Table~\ref{tab:auth_comp} (bottom half) presents examples of completions for longer and more complex utterances (three words or more,
see Supplementary Materials for more examples).
For such longer utterances, the models demonstrate a better grasp of context, producing more complete and semantically coherent responses, with less redundancy and fewer nonsensical completions. 
Even for uncommon phrases, the models are better at generating plausible continuations. For instance, ``Oh is it hid feet'' was completed to ``Oh is it hid under the feet'' by \textsc{T5-base}, which, despite a minor grammatical error, shows an effort to logically complete the utterance.

\section{Discussion}
\label{sec:Discussion}

In this study, we aimed to explore the potential of reconstructing Broca’s aphasic language into full utterances using LLMs fine-tuned on synthetic Broca’s aphasic data. Using a rule-based system, we generated synthetic data based on transcribed spoken language from neurotypical individuals. Our synthetic data mirrors the characteristics of Broca's aphasic speech. Subsequently, we fine-tuned four LLMs on this synthetic data to evaluate their ability to complete agrammatic aphasic sentences. To the best of our knowledge, this is the first study to evaluate LLMs on the task of reconstructing authentic Broca's aphasic sentences.

While beneficial, current rehabilitation methods for Broca's aphasia often fall short in the long-term sustainability of treatment, demand considerable time and resources and do not fully cover the diverse communication challenges faced by PWA. 
LLM-based approaches provide a possibility to overcome these restraints by offering ongoing and consistent support tailored to the specific needs of a patient, improving their quality of life and that of their spouses and caregivers.

\subsection{Automatic agrammatic sentence completion}

\subsubsection{Synthetic data}

We found that the models performed well on synthetic aphasic data, generating coherent and contextually appropriate completions. The \textsc{T5-base} and \textsc{Flan}-T5-XL models showed consistent performance in the synthetic setting, with \textsc{Flan}-T5-XL exhibiting a slightly better use of context and producing slightly better completions.
These findings are consistent with the study of\cite{misra2022assistive}, who demonstrated the utility of synthetic data for training LLMs to reconstruct agrammatic aphasic sentences. However, while \cite{misra2022assistive} used written text from the C4 corpus to generate synthetic aphasic utterances, our approach is based on spoken language, which better reflects conversational contexts encountered by PWA. Additionally, unlike \cite{misra2022assistive} who evaluated their model solely on synthetic data, we extended the evaluation to authentic aphasic sentences, addressing a key limitation in their study. This broader evaluation provides a more realistic assessment of the models’ applicability to clinical scenarios.

\subsubsection{Authentic data}
When tested on authentic Broca's aphasic sentences, the models' performance varied. The models were able to generate reasonable completions based on visual inspections, with better results observed for longer input sentences, which provide more contextual information, resulting in improved model performance. Visual inspections indicated that \textsc{Flan}-T5-XL generated more coherent and contextually relevant completions compared to \textsc{T5-base}, particularly for longer utterances. 
In contrast to \cite{misra2022assistive} study, which did not test their approach on real aphasic data, our findings demonstrate that fine-tuned LLMs can generalize effectively to authentic agrammatic sentences produced by individuals with Broca’s aphasia. This step highlights the potential and the practical relevance of such models for rehabilitation purposes. While previous AI applications in aphasia have largely focused on diagnostic tasks or static assistive technologies \cite{Adikari2024, Privitera2024, Azevedo2024, azevedo2023understanding}, our results suggest that LLM-based systems have the potential to provide adaptive solutions for reconstructing agrammatic speech into complete sentences. Incorporating longer input utterances further improves model performance, suggesting the importance of designing systems capable of leveraging contextual information to support real-world communication needs for PWA and their caregivers.

\subsubsection{Implications}

The demonstrated ability of sequence-to-sequence LLMs to generate reasonable completions for Broca's aphasic sentences highlights their potential in advancing communicative technologies for non-fluent aphasic patients. These models could become valuable tools in speech and language therapy by providing contextually appropriate sentence completions, potentially helping individuals with Broca's aphasia convey their thoughts more effectively.

Moreover, the study emphasises the importance of leveraging authentic data to build robust models for real-world applications. The performance improvements observed when focusing on longer utterances imply that taking into account a minimum input length could enhance the effectiveness of such models in practical settings. 

Beyond the application to Broca's aphasia, the methodology used for generating synthetic data could be expanded, in order to build a comprehensive database of synthetic speech and language data not only for individuals with Broca's aphasia but also for other types of aphasia and potentially for other clinical populations with speech and language disorders. 

Synthetic data also addresses ethical concerns such as patient privacy and consent, providing an anonymous alternative. It provides a cost-effective way to create large datasets, which reduces the need for expensive, time-consuming data collection \cite{melamud-shivade-2019-towards}, although it is important to keep in mind that no synthetic data can completely replace authentic human data.

\subsection{Limitations} \label{sec:limitations}

The study's focus on Broca's aphasia possibly limits the generalisability of the findings to other types of non-fluent aphasia or related language impairments, since different forms of non-fluent aphasia present distinct linguistic challenges. The study also does not fully address the variability in individual language use and recovery patterns among people with Broca's aphasia, leading to suboptimal performance for certain users. 

The computational resources required for training and fine-tuning large transformer-based models are substantial. This limitation may restrict the accessibility and scalability of the proposed approaches, specifically for researchers and practitioners with limited computational resources. Moreover, we only trained and tested the models on one high-resource language, English, and future work should determine whether our proposed method can also be used in other languages.

Finally, recovering the intended meaning of agrammatic aphasic sentences without extensive context or direct interaction with the individual is inherently challenging. For instance, the fragment ``He book table'' could be interpreted in multiple ways such as ``He put the book on the table'', ``He took the book from the table'' or ``He wants the book on the table.'' This variability emphasises the challenge that a sentence cannot always be accurately recovered without additional grounding such as understanding the broader conversation or using external data such as visual cues.

\subsection{Future work} \label{sec:future_work}

In terms of synthetic Broca's aphasic data generation, one route for future research is the exploration of data augmentation techniques tailored to agrammatic aphasic sentences, improving generalisation capabilities of the models on real-world Broca's aphasic speech.

Another interesting direction for future work is studying the domain influence, possibly enabling the creation of task-specific models based on the patient's interests or hobbies.
The incorporation of domain knowledge has demonstrated improved model performance in NLP, including machine translation \cite{khiu-etal-2024-predicting} and NLP-based educational applications \cite{sakakini-etal-2019-equipping}. 
In terms of training setup, it could be promising to explore in-context learning methods with the same models, and compare the resulting scores with those reported in our study (similar, to, e.g., \cite{mosbach2023few}). Our preliminary experiments with FLAN-T5-XL in a ten-shot learning setup suggest that fine-tuning methods yield better performance (on ChrF and Rouge), but experimenting with a variety of prompts may be necessary in the future.

Finally, we recognize that the observed correct sentence completion rate is not sufficient for deployment as a ready-to-use communication aid in clinical settings. Nonetheless, this work represents an initial step toward building such applications, since there are currently no existing benchmarks for performance on authentic aphasic sentences. Therefore, future research could ask native English speakers to produce corresponding target completions for the authentic Broca's aphasic utterances from AphasiaBank. The availability of such ground-truth data is crucial for a quantitative evaluation of models' completions of the authentic Broca's aphasic sentences. The creation of a reliable benchmark would provide a reference for evaluating how well the sentence completion models handle typical errors and omissions in Broca's aphasia. Moving beyond the sentence completion task, such data would contribute immensely to creating a broader benchmark for related tasks in the field of aphasiology.

\section{Conclusion}
\label{sec:Conclusion}

In summary, we conclude that sequence-to-sequence large language models exhibit the capability of reconstructing agrammatic sentences produced by individuals with Broca's aphasia. The models generate reasonable and contextually appropriate completions, denoting their potential as valuable tools in speech and language therapy. These findings support the ongoing development and refinement of machine learning models tailored to assist in communication for those affected by language impairments. The broader potential to create synthetic data for various clinical populations opens avenues for improving their communicative independence and quality of life.

\bibliography{bibliography}

\onecolumn

\section{Supplementary Materials}

\subsection{BLEU exclusion}
\label{apx:bleu_exclusion}

Since the sentence completion task is inherently different from machine translation, we do not use the conventional BLEU \cite{papineni2002bleu} metric. 
BLEU often results in scores of 0.00 for reasonable completions because it relies on exact n-gram matches, which can be too strict for this task. Even slight variations in word choice or order can lead to a BLEU score of 0.00 if the n-grams do not align perfectly.
Table \ref{tab:ok_completions} shows a few examples of reasonable completions assigned a BLEU score of 0.00 along with the assigned ChrF and RougeL scores, demonstrating the appropriateness of ChrF and RougeL over BLEU for this task. ChrF considers character-level n-grams, providing a more nuanced evaluation by capturing partial matches and minor variations. RougeL finds the longest common subsequence between the generated and reference text, allowing us to estimate how well the generated sentence preserves word order and grammatical coherence. Thus, offering a better assessment of the quality of the generated completions. Moreover, Table \ref{tab:bleu_stats_flant5xl_xts} shows the descriptive statistics of the BLEU metric calculated on all completions of the synthetic test set by FLAN-T5-XL fine-tuned using the prefix, further indicating the unsuitability of the BLEU metric for the sentence completion task. 

\begin{table}[h]
\caption{Examples of reasonable completions with BLEU scores of 0.00 and their corresponding ChrF and RougeL scores. All examples are taken from FLAN-T5-XL fine-tuned using the prefix.}
\label{tab:ok_completions}
\resizebox{\columnwidth}{!}{\begin{tabular}{lccccc}
\toprule
Syn.\ utterance    & Target completion   & Gen.\ completion   & BLEU score  & ChrF score & RougeL score   \\ \midrule
And run it computer.       & And run it through the computer.      & And run it on the computer.  & 0.00   & 63.75 &  0.83 \\ 
Janine uses Charlie times.  & Janine uses Charlie all the time.    & Janine uses Charlie several times.   & 0.00    & 65.45 & 0.55 \\ 
Without the risk of killed.      & Without running the risk of getting killed.       & Without the risk of being killed.   & 0.00    & 56.87 &  0.77 \\ 
Gradually bring it.   &   Just gradually bring it around.   &  Gradually bring it up.   & 0.00  & 55.65   & 0.44 \\  
And then they.   &   And then they break.   &  And then they sat down.   & 0.00  & 58.39   & 0.67 \\ 
\bottomrule 
\end{tabular}}
\end{table}

\begin{table}[h!]
\centering
\caption{Descriptive statistics of the BLEU score computed for FLAN-T5-XL on the synthetic test set using the prefix ``Complete this sentence''.}
\label{tab:bleu_stats_flant5xl_xts}
\begin{tabular}{cc}
\toprule
Statistic              & Score   \\
\midrule
Mean                 & 0.24    \\
Standard deviation   & 0.30    \\
Min                  & 0.00    \\
Median               & 0.00   \\
\bottomrule
\end{tabular}
\end{table}

\subsection{Fine-tuning setup details}
\label{app:finetuning}
During fine-tuning, we initialise the QLoRA configuration (see Tables below for quantisation parameters and hyperparameters), prepare the model for 4-bit fine-tuning and add the LoRA adaptor using the initialised configuration.

\begin{table}[h!]
\centering
\begin{minipage}{0.45\textwidth}
    \centering
    \caption{Quantisation parameters in the BitsAndBytes configuration.}
    \label{tab:bitsAndBytes}
    \begin{tabular}{l}
    \toprule
    BitsAndBytes configuration parameters  \\ \midrule
    load\_in\_4bit=True                                   \\ 
    bnb\_4bit\_use\_double\_quant=True                        \\ 
    bnb\_4bit\_quant\_type=``nf4''                    \\ 
    bnb\_4bit\_compute\_dtype=torch.bfloat16                 \\ 
    \bottomrule
    \end{tabular}
\end{minipage}
\hspace{0.05\textwidth}
\begin{minipage}{0.45\textwidth}
    \centering
    \caption{Hyperparameters in the LoRA configuration.}
    \label{tab:paramset}
    \begin{tabular}{l}
    \toprule
    LoRA configuration parameters  \\ \midrule
    r=8                                    \\ 
    lora\_alpha=32                         \\ 
    lora\_dropout=0.05                     \\ 
    inference\_mode=False                  \\
    bias=``none''                            \\
    task\_type=TaskType.SEQ\_2\_SEQ\_LM    \\ 
    \bottomrule
    \end{tabular}
\end{minipage}
\end{table}

\subsection{Training object}
\label{apx:train_args}

We use the Seq2SeqTrainingArguments and Seq2SeqTrainer classes from the HuggingFace transformers library \cite{wolf-etal-2020-transformers} to define the training arguments and trainer object respectively.

\begin{table}[h!] 
\centering
\caption{The training arguments (Seq2SeqTrainingArguments) used to fine-tune the models.}
\begin{tabular}{lll}
\toprule
Hyperparameter                  & Value for T5-base  & Value for all other models \\ \midrule
output\_dir                     & checkpoints\_path & checkpoints\_path \\ 
learning\_rate                  & 1e-4 & 1e-4 \\ 
per\_device\_train\_batch\_size & batch\_size & batch\_size \\ 
per\_device\_eval\_batch\_size  & batch\_size & batch\_size \\ 
num\_train\_epochs              & \textbf{5} & \textbf{1} \\ 
weight\_decay                   & 0.01 & 0.01 \\ 
predict\_with\_generate         & True & True\\ 
evaluation\_strategy            & ``epoch'' & ``epoch''\\ 
save\_strategy                  & ``epoch'' & ``epoch'' \\ 
load\_best\_model\_at\_end      & True & True \\ 
fp16                            & True & True \\ 
gradient\_accumulation\_steps   & 2 & 2 \\
logging\_dir                    & checkpoints\_path + ``/logs'' & checkpoints\_path + ``/logs'' \\ \bottomrule
\end{tabular}
\label{apx:tab_train_args_}
\end{table}

\begin{table}[h] 
\centering
\caption{The trainer object definement (Seq2SeqTrainer) used to fine-tune the sentence completion models.}
\begin{tabular}{ll}
\toprule
Hyperparameter   & Value \\ \midrule
model            & model \\ 
args             & training\_args \\ 
train\_dataset   & tokenized\_train\_dataset \\ 
eval\_dataset    & tokenized\_valid\_dataset \\ 
tokenizer        & tokenizer \\ 
data\_collator   & data\_collator \\ 
compute\_metrics & compute\_metrics (function) \\ 
callbacks        & [EarlyStoppingCallback(early\_stopping\_patience=2)] \\ \bottomrule
\end{tabular}
\label{apx:tab_train_obj_}
\end{table}

\FloatBarrier

\subsection{Data pre-processing}
\label{app:preprocessing}

\begin{table}[h]
\caption{Examples of unprocessed and pre-processed data obtained from AphasiaBank and SBCSAE in CHAT format.}
\label{tab:processed_data}
\resizebox{\columnwidth}{!}{
\begin{tabular}{l|cc}
\toprule
Corpus & Unprocessed text & Pre-processed \\ \midrule
AphasiaBank (Aphasic) & it's alright . [+ exc] 360360\_360780 & It is alright. \\
AphasiaBank (Aphasic) & \&=imit:running and goes +"/. 453922\_454482 & And goes. \\
AphasiaBank (Aphasic) & +"" \&+b I wanna come . 727610\_728460 & I want to come. \\
AphasiaBank (Aphasic) & that he [/] he said . 804035\_805805  & That he said. \\ \midrule
SBCSAE & (..) So you don't need to go (..) borrow equipment from anybody 0\_9210  to +/. & So you do not need to go borrow equipment from anybody to. \\ 
SBCSAE &  \&=in \&=lengthened (.) and (..) \&=tsk (..) I don't know . 76950\_79570 & And i do not know. \\
SBCSAE & you know 231970\_232310 and you'd put the +/. 232310\_232940 & You know and you would put the. \\ 
SBCSAE & \&=in you'd have +/. 232940\_233690 & You would have. \\ \bottomrule
\end{tabular}
}
\end{table}

The pre-processing steps include replacing multiple consecutive spaces by a single space, unfolding contractions using the the contractions library \cite{contractions} and the removal of the following features:
\begin{itemize}
    \item Instances of trailing off pauses such as ``(..)'', ``(...)'' and ``+...'';
    \item Filler pauses including ``\&-um'', ``\&-uh'', ``\&-mm'' and ``\&-like'';
    \item Actions described within the text, indicated by sequences such as ``\&=imit:runnin'';
    \item All Unicode errors present in the text;
    \item Sequences beginning with two special characters followed by text;
    \item Text within square brackets ``[]'' and angle brackets ``<>'';
    \item Strings starting with ``\&+'' or ``U'' followed by text as well as those beginning with characters such as ``*'', ``?'' and ``!'';
    \item Arbitrary length strings of Xs;
    \item Strings starting with special characters followed by one or more numbers;
    \item Any remaining special characters such as ``‡'', ``+/.'', ``@o'' and ``=@'';
    \item Instances of stuttering and repetitive words.
\end{itemize}
The effect of these pre-processing steps can be observed in Table \ref{tab:processed_data}, which presents examples of the data before and after pre-processing. This cleaning process ensures that the resulting dataset is as free from noise and inconsistencies as possible.

\newpage
\FloatBarrier

\subsection{Synthetic example completions} \label{apx:syn_exa_comp}

\begin{table}[h]
\centering
\caption{Additional examples of generated completions on the synthetic test set.}
\label{tab:syn_comp_exa}
\begin{tabular}{l|cc}
\toprule
Model    & Synthetic aphasic utterance & Generated completion     \\ \midrule
T5-base-cts     & It those i went out and coffees.      & It was all those i went out and got coffee.  \\
T5-base-cts     & They been it a while.      & They have been doing it for a while.  \\
T5-base-cts     & Person meaning who.     & Person meaning who.  \\  
T5-base-cts     & Make it like we are it.      & Make it look like we are going to do it.   \\  
T5-base-cts     & They so fast they shape containers.      & They are so fast they can shape containers.  \\
T5-base-cts     & And i called talk him.      & And i called to talk to him.  \\ 
T5-base-cts     & Well top guys lip.      & Well top guys lip.  \\  
T5-base-cts     & Oh i it they.      & Oh i thought it was they.   \\ 
T5-base-cts     & I see venice high school.      & I see venice high school.   \\ 
T5-base-cts     & It great in rooms we.      & It was great in the room we were in. \\ \midrule 
\textsc{Flan}-T5-XL-cts  & And they control.      & And they are in control.  \\
\textsc{Flan}-T5-XL-cts  & I a lot      & I have a lot of work to do.  \\
\textsc{Flan}-T5-XL-cts  & I a rake but i it.      & I have a rake but i don't use it.   \\ 
\textsc{Flan}-T5-XL-cts  & There an orders to how i it.      & There is an order to how i want to do it. \\ 
\textsc{Flan}-T5-XL-cts  & I worked my money.      & I worked hard for my money.  \\
\textsc{Flan}-T5-XL-cts  & I he going down.      & I think he is going down.   \\
\textsc{Flan}-T5-XL-cts  & This is you.      & This is for you.  \\
\textsc{Flan}-T5-XL-cts  & You are a aheads me son.      & You are going to be ahead of me son.  \\ 
\textsc{Flan}-T5-XL-cts  & And he good at making out budget.      & And he is good at making out a budget.   \\
\textsc{Flan}-T5-XL-cts  & Oh he is be you know closet everything.      & Oh he is going to be you know in the closet with everything. \\ \bottomrule
\end{tabular}
\end{table}

\FloatBarrier

\subsection{Authentic example completions} \label{apx:auth_exa_comp}

\begin{table}[h]
\centering
\caption{Additional examples of generated completions for all the authentic data. The prefix is used in all cases.}
\label{tab:auth_comp_exa}
\resizebox{\columnwidth}{!}{
\begin{tabular}{l|cc}
\toprule
Authentic aphasic utterance   & Generated by T5-base            & Generated by \textsc{Flan}-T5-XL     \\
\midrule
A dog chase a up to a tree.        & A dog chases a dog up to a tree.      &  A dog chases a ball up to a tree.  \\                   
Leg shirt leg pants leg.          & Leg shirt leg pants leg.      &  Leg shirt leg pants leg.  \\ 
I was think of going yeah.   & I was thinking of going yeah.      & I was thinking of going to the store yeah.  \\ 
High that was a the firetruck.   & High that was a firetruck.      &  High that was a firetruck.  \\ 
Alright fine.     & Alright fine.     &  Alright fine.  \\  
The prince the.   & The prince of the.      &  The prince is the. \\  
And get some the bread.   & And get some of the bread.      &  And get some of the bread.  \\ 
I did not know words. & I did not know how to say.   &   I did not know what the words were.    \\ 
And i did not even what i was wrong.  & And i did not even know what i was doing wrong.   &    And i did not even know what i was doing was wrong.    \\ 
Well i do not have to do that.      & Well i do not have to do that.   &  Well i do not have to do that.  \\ 
My firstnamer my daughter.     & My firstname is my daughter.   &      My firstname is my daughter.     \\ 
But i can listen to the books.     & But i can listen to the books.    &    But i can listen to the books.      \\ 
Okay they will tell me restaurants.    & Okay they will tell me about the restaurant.   &      Okay they will tell me about the restaurants.       \\ 
She tells me lotsof words.         & She tells me lots of words.     &    She tells me lotsof words.   \\ 
And then about six sixth years i called her.   & And then about six sixth years later i called her.   &   And then about six or six and a half years later i called her.  \\ \bottomrule
\end{tabular}
}
\end{table}

\begin{table}[H]
\centering
\caption{Additional examples of generated completions for the authentic data with one-word and two-word utterances removed.}
\label{tab:auth_comp_3plus_exa}
\begin{tabular}{l|cc}
\toprule
Authentic aphasic utterance & T5-base-cts completion & \textsc{Flan}-T5-XL-cts completion \\ \midrule
The this one ?. & Then the next one? & Then this is the one?. \\ 
Oh it is ?. & Oh it is ok? & Oh it is not?. \\ 
And they went kiss and go. & And they went to kiss and go. & And they went to kiss and go.  \\ 
And there's one two. & And there's one of them. & And there's one or two. \\ 
Then the this andjelly over that. & Then the next time the andjelly over that. & Then the jar is put on this andjelly over that.  \\ 
You in er in. & You are in er in. & You are in er in.  \\ 
And that is that is the worst part. & And that is that is the worst part. & And that is that is the worst part.  \\ 
It is one major youknow wne. & It is one major youknow wn. & It is one major youknow wn.  \\ 
Imean that is that is just. & Imean that is that is just. & I mean that is just a little bit.  \\ 
Stroke numb youknow i burst. & Stroke is numb you know i burst. & Stroke numb youknow i burst.  \\ 
So mary'shospital surgery. & So mary's hospital surgery. & So mary'shospital had surgery.  \\ 
Friend support cousin family. & Friend support cousin family. & Friend support cousin family.  \\ 
Say beautiful so manage. & Say beautiful so manageable. & Say beautiful so you can manage.  \\ 
So finally police. & So finally the police. & So finally the police.  \\
Well it two am stroke. & Well it was two am stroke. & Well it is two am having a stroke. \\ \bottomrule  
\end{tabular}
\end{table}

\end{document}